# Optimal Bandwidth Selection for DENCLUE Algorithm

Hao Wang
CEO Office
Ratidar Technologies LLC
*Beijing, China*
haow85@live.com

*Abstract*—In modern day industry, clustering algorithms are daily routines of algorithm engineers. Although clustering algorithms experienced rapid growth before 2010. Innovation related to the research topic has stagnated after deep learning became the de facto industrial standard for machine learning applications. In 2007, a density-based clustering algorithm named DENCLUE was invented to solve clustering problem for nonlinear data structures. However, its parameter selection problem was largely neglected until 2011. In this paper, we propose a new approach to compute the optimal parameters for the DENCLUE algorithm, and discuss its performance in the experiment section.

## I. INTRODUCTION

Clustering algorithms are routines in the internet companies. It is impossible to talk about machine learning without having grasped basic knowledge of clustering techniques. As a class of unsupervised learning algorithms, clustering approaches have drawn a large number of researchers prior to the emergence of supervised learning in 1990's. The field has stagnated compared to new technologies such as deep learning in more recent times.

Clustering is important because it is capable of solving many real world problems. For example, in bio-informatics, we need to find classification of species, but we do not have training data. In this case, clustering is an appropriate technique that is versatile and powerful.

Clustering algorithms can be classified into many different categories. One such category is density-based clustering algorithms. Density-based clustering algrithms include DBSCAN [1], DENCLUE [2], etc. DENCLUE was invented in 2007. The algorithm relies heavily on non-parametric statistics, especially Nadaraya-Watson Kernel Estimator [3]. However, DENCLUE needs to choose bandwidths for its computation. The bandwidth selection problem was not tackled until the year of 2011 [4].

Density-based clustering algorithm has great advantages over other types of clustering algorithms because it could classify complex shapes that are not linearly separable.

## II. RELATED WORK

Clustering algorithm is a set of machine learning algorithms that are capable of classifying datasets without reference to training datasets. Instead of relying on supervised learning methodologies, clustering algorithms resort only to parameter selection to tackle the problem of machine learning.

One of the most notable inventions in the field is the K-Means clustering algoirthm [5] and its variants such as K-Medoids [6] and Fuzzy K-Means [7]. A more recent invention related K-Means clustering uses non-parametric statistics to solve the problem [8].

A different school of thoughts in the field of clustering algorithm research is density-based clustering algorithms. One classic example of the category is DBSCAN [1], which could solve non-linearity problems. Another great example is DENCLUE [2], which uses non-parametric statistics to attack the clustering problem.

Optimal bandwidth selection approaches for non-parametric models are popular research topics between 1960's and 2000's. Important work include optimal bandwidth selection for kernel regression [9] [10], kernel density estimation [11][12]. Similar ideas are also carried over to the field of parametric models such as logistic regression [13][14]. In this paper, we follow the line of research thoughts to solve the optimal parameter selection problem for the DENCLUE algorithm. DENCLUE has several variants and parameter selection techniqeus that are also worth noticing [15][16].

## III. DENCLUE ALGORITHM

The DENCLUE algorithm is a density-based clustering algorithm that uses technologies such as Nadaraya-Watson Kernel Estimator. The algorithm, in precise terms, belongs to the DBSCAN family. The algorithm needs to find the local maxima using the Nadaraya-Watson Kernel Estimator, which is solved using a hill-climing approach. The points within a certain distance range are considered as belonging to the same cluster of the local maximum. The points are also assigned a very small probability as being an outlier. In one of the research publications, the problem is reduced to an EM algorithm.

To save the space for this research publication, we omit the details of the mathematical formulas for the DENCLUE algorithm in this section. Interested readers could check [2] for detailed introduction of the algorithm.

## IV. OPTIMAL BANDWIDTH SELECTOR FOR DENCLUE ALGORITHM

The optimal bandwidth selector for DENCLUE algorithm's hill climbing procedure is to solve the following problem :

$$\operatorname*{argmax}_h \text{MSE} = \|x_i - \sum_{j=1}^{N} k_h(\frac{x_j - x_i}{h}) x_j\|^2$$

We take a particualr example of the Nadaraya-Watson Kernel Estimator, namely Gaussian Kernel, and apply Stochastic Gradient Descent (SGD) algorithm to solve the problem. We obtain the following formula (M is the dimension of the data points) :

$$h = h + \eta(2Mt_0 t_1 h^{-(M+1)}(x - t_1 p)^T \cdot p)$$

$$t_0 = \|x - p\|_2^2$$

$$t_1 = exp(-t_0 h^{-M})$$

The update rule for the computation of h can be easily implemented into the full-functioning code of DENCLUE algorithm.

## V. EXPERIMENTS

We test our algorithms on the following 2 datasets, using gradient learning step value $3 \times 10^{-3}$, and visualize the experimental results in the following plots :

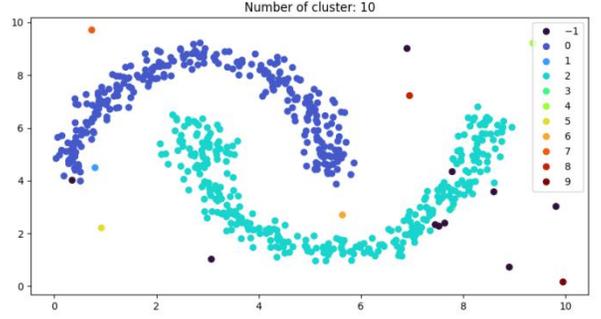

Fig. 1 Clustering Example on Test Data 1

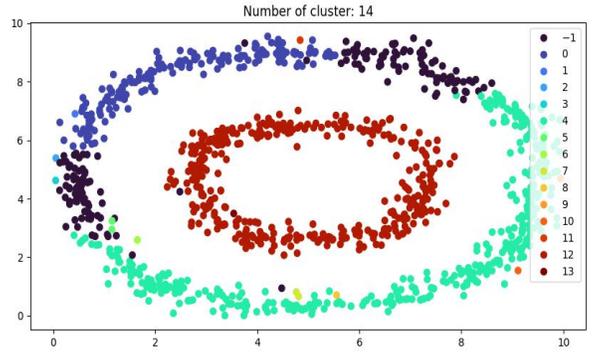

Fig. 2 Clustering Example on Test Data 2

From Fig. 1 and Fig. 2, we notice that DENCLUE with the optimal bandwidth selector could classify point data pretty accurately for Fig. 1, but generate more clusters than our intuitive perception for Fig. 2 (Class Label -1 means outliers).

Changing the gradient learning step value to $6 \times 10^{-3}$ , we obtain the following plots :

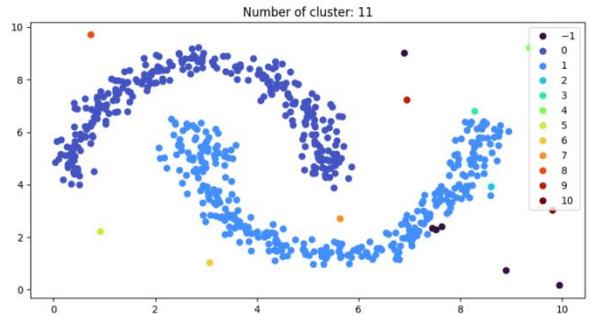

Fig. 3 Clustering Example on Test Data 1

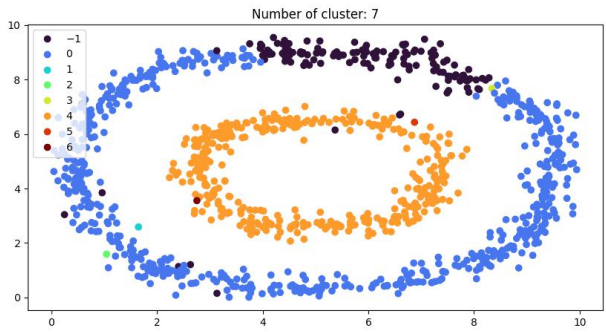

Fig. 4 Clustering Example on Test Data 2

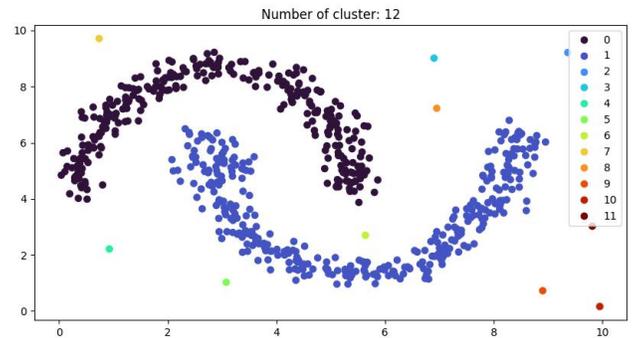

Fig. 5 Clustering Example on Test Data 1

Fig. 3 shows that the data set is classified pretty accurately. Fig. 4 shows that the data set is classified fairly accurately, except that a section of the point data is classified as outliers. We believe by tweaking the gradient learning step values, we will be able to generate better and better experimental results.

From the above experiments, we realized that the outlier probability threshold is too large, and after making it smaller, we obtain Fig. 5 and Fig.6, the dataset in which looks just perfectly classified.

## VI. Conclusion

In this paper, we proposed an algorithm to compute the optimal bandwidth for the DENCLUE algorithm. The algorithm uses Stochastic Gradient Descent (SGD) to approximate the ground truth optimal values of the algorithm. The experimental results show that the algorithm with optimal parameters can classify non-linearly separable datasets.

In future work, we would like to explore other nonparametric statistical techniques to solve machine learning including clustering algorithms. We would also like to find parameter-free solutions to the aforementioned problems. Another issue that bugs us is the speed of DENCLUE algorithm using our optimal bandwidth selector. We will imporve the speed of our algorithm in future work.

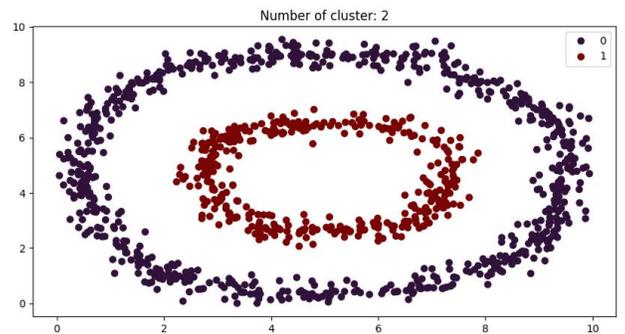

Fig. 6 Clustering Example on Test Data 2